\def\year{2021}\relax
\documentclass[letterpaper]{article} % DO NOT CHANGE THIS
\usepackage{aaai21}  % DO NOT CHANGE THIS
\usepackage{times}  % DO NOT CHANGE THIS
\usepackage{helvet} % DO NOT CHANGE THIS
\usepackage{courier}  % DO NOT CHANGE THIS
\usepackage[hyphens]{url}  % DO NOT CHANGE THIS
\usepackage{graphicx} % DO NOT CHANGE THIS
\usepackage{natbib}  % DO NOT CHANGE THIS AND DO NOT ADD ANY OPTIONS TO IT
\usepackage{caption} % DO NOT CHANGE THIS AND DO NOT ADD ANY OPTIONS TO IT
\usepackage{times}

\usepackage{subfiles}
\usepackage[bookmarks=true]{hyperref}
\usepackage{balance}
\usepackage{hyperref}
\usepackage{graphicx}
\usepackage{amssymb}
\usepackage{amsmath}
\usepackage{amsfonts}
\usepackage{caption}
\usepackage{algorithm}
\usepackage{algorithmicx}
\usepackage[noend]{algpseudocode}
\usepackage{graphicx,caption}
\usepackage{amsmath,amssymb,stmaryrd,mathtools}
\usepackage{url}
\usepackage{subcaption}
\usepackage{booktabs}
\usepackage{multirow}
\usepackage{float}
\usepackage{amsmath}
\usepackage{amsthm}
\usepackage{adjustbox}
\DeclareMathOperator*{\argmax}{arg\,max}

\hypersetup{
    colorlinks=true,
    linkcolor=blue,
    citecolor=blue,
    filecolor=magenta,      
    urlcolor=cyan,
    hyperindex=True,
 urlcolor=blue}

\usepackage{wrapfig}

\newcommand{\Abhi}[1]{}

\usepackage{soul}
\newcommand{\expect}[2]{\mathbb{E}_{#1}\left[#2\right]}

\newcommand{\ldeltafunc}[1]{\mathbf{1}_{\left[#1\right]}}
\newcommand{\logp}[1]{\log\left(#1\right)}

\newcommand{\defeq}{\vcentcolon=}

\renewcommand{\d}{\mathop{}\!\mathrm{d}}
\usepackage{amsmath,amssymb,stmaryrd,mathtools}

\renewcommand{\vec}[1]{\mathbf{#1}}
\newcommand{\x}{\vec{x}}

\newcommand{\X}{\vec{X}}
\newcommand{\OX}{\Omega_\X}

\newcommand{\z}{\vec{z}}

\newcommand{\Z}{\vec{Z}}

\newcommand{\OZ}{\Omega_\Z}

\hypersetup{
    colorlinks=true,
    linkcolor=blue,
    citecolor=blue,
    filecolor=magenta,      
    urlcolor=cyan,
    hyperindex=True,
 urlcolor=blue}
\title{Generalized Adversarially Learned Inference}
\author{
   Yatin Dandi$^1$, Homanga Bharadhwaj$^2$, Abhishek Kumar$^3$, Piyush Rai$^1$
}
\affiliations{
    $^1$Indian Institute of Technology Kanpur
    \\ $^2$University of Toronto, Vector Institute
    \\ $^3$Google Brain
}
\begin{document}

\maketitle

\begin{abstract}
Allowing effective inference of latent vectors while training GANs can greatly increase their applicability in various downstream tasks. Recent approaches, such as ALI and BiGAN frameworks, develop methods of inference of latent variables in GANs by adversarially training an image generator along with an encoder to match two joint distributions of image and latent vector pairs. We generalize these approaches to incorporate \textit{multiple} layers of feedback on \textbf{reconstructions}, \textbf{self-supervision}, and other forms of supervision based on prior or \textbf{learned knowledge} about the desired solutions. We achieve this by modifying the discriminator's objective to correctly identify more than two joint distributions of tuples of an arbitrary number of random variables consisting of images, latent vectors, and other variables generated through auxiliary tasks, such as reconstruction and inpainting or as outputs of suitable pre-trained models. We design a non-saturating maximization objective for the generator-encoder pair and prove that the resulting adversarial game corresponds to a global optimum that simultaneously matches all the distributions. Within our proposed framework, we introduce a novel set of techniques for providing self-supervised feedback to the model based on properties, such as patch-level correspondence and cycle consistency of reconstructions. Through comprehensive experiments, we demonstrate the efficacy, scalability, and flexibility of the proposed approach for a variety of tasks.
\end{abstract}

\section{Introduction}
Recent advances in deep generative models have enabled modeling of complex high-dimensional datasets. In particular, Generative Adversarial Networks (GANs) \cite{gan} and Variational Autoencoders (VAEs) \cite{vae} are broad classes of current state-of-the-art deep generative approaches, providing complementary benefits. VAE based approaches aim to learn an explicit inference function through an encoder neural network that maps from the data distribution to a latent space distribution. On the other hand, GAN based adversarial learning techniques do not perform inference and directly learn a generative model to construct high-quality data, which are usually much more realistic than those generated by VAEs. However, due to the absence of an efficient inference mechanism it is not possible to learn rich unsupervised feature representations from data.
\begin{figure*}\begin{minipage}[b]{.65\textwidth}\centering \includegraphics[width=\textwidth]{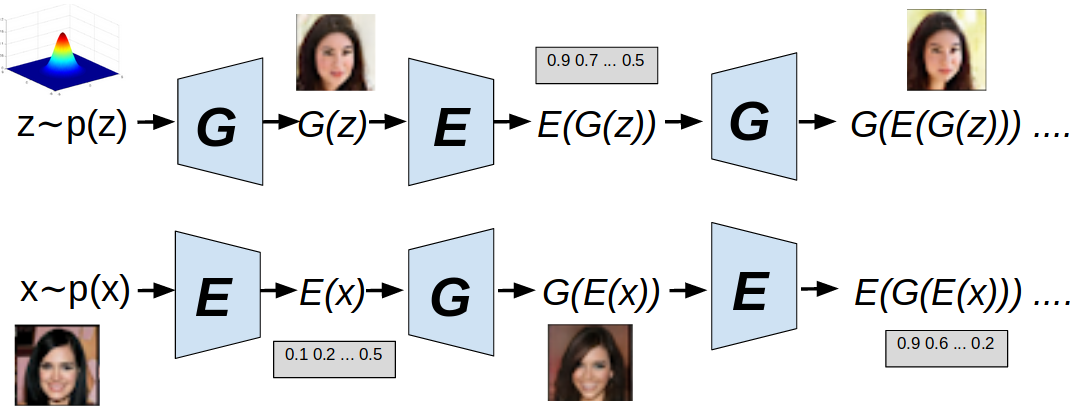}\subcaption{Chain of reconstructive feedback. The images denote real reconstructions from GALI-4.}\label{fig:mainfig}\end{minipage}%
\hfill
\begin{minipage}[b]{.35\textwidth}\centering \includegraphics[width=\textwidth]{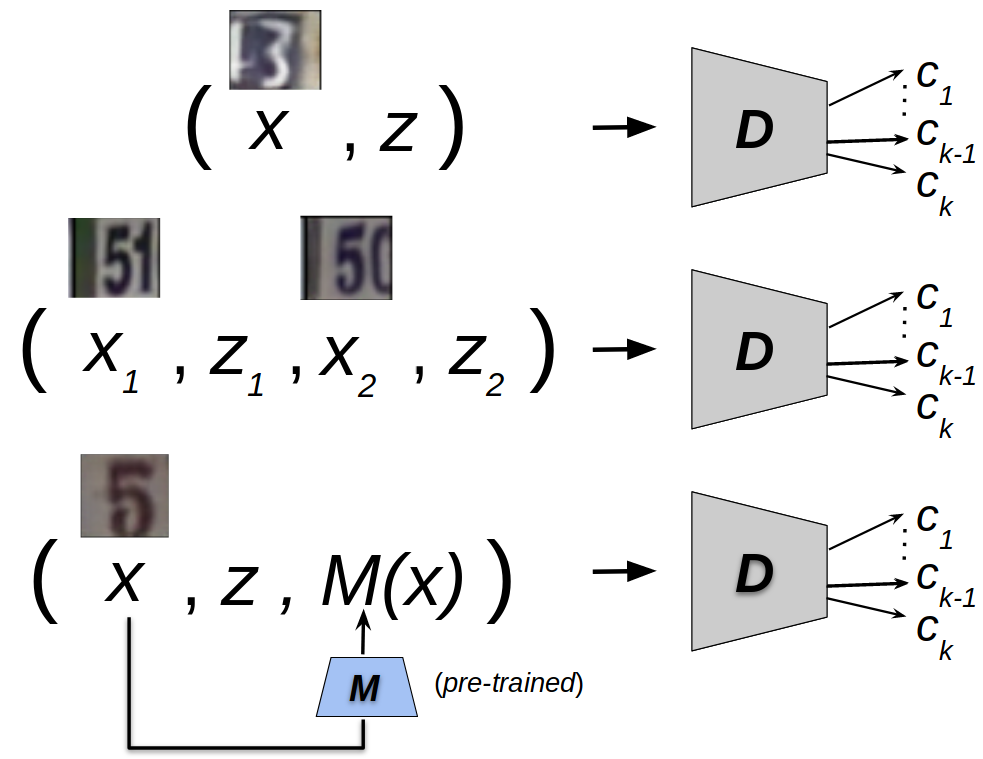}\subcaption{Different tuple types}\label{fig:tuples}\end{minipage}
\caption{(a) The two sequences generated by recursively applying the encoder and the generator to images and latent vectors respectively. The top sequence starts from a latent vector drawn from the fixed prior while the bottom sequence starts from a real image. The images correspond to actual reconstructions from the proposed GALI approach. In particular, GALI-4 (i.e. GALI with $n=4$ as described in "The Proposed Approach" was used to generate the reconstructions. In the figure $G$ denotes the generator, $E$ the encoder, $z$ the latent vector, and $x$ the input image. (b) Illustration of the different tuple types that can be input to the discriminator to provide different types of feedbacks via distribution matching. Here, $M$ denotes an external pre-trained neural network and $M(x)$ denotes the features corresponding to image $x$ through model $M$. The multi-class discriminator $D$ outputs one of the $k$ classes.}\label{fig:main}
\vspace{-1em}
\end{figure*}

To address the above issues, recent approaches, in particular, Adversarially Learned Inference (ALI) and Bidirectional GAN (BiGAN)~\cite{ali,bigan} have attempted to integrate an inference mechanism within the GAN framework by training a discriminator to discriminate not only in the data space ($x$ vs $G(z)$), but discriminate \emph{joint} samples of data and encodings ($(x,E(x))$) from joint samples of the generations and latent variables ($(G(z),z)$). Here, $E(\cdot)$ denotes the encoder and $G(\cdot)$ denotes the generator. %(or Generator as per the terminology in GAN).% %Technically, both $E(\cdot)$ and $G(\cdot)$ are generators w.r.t. the adversarial games described in ALI~\cite{ali} and BiGAN~\cite{bigan} because they represent \textit{different transformation processes}: from the data to latent representations, and from the latent representation to data, respectively% 
%\abhi{I am not sure if I follow this argument: normally the term ``generative process'' is used to refer to generative mechanism of observed data only}. \YD{maybe we could say that the generation and inference processes both play identical roles with respect to the discriminator as different transformation processes?}
We argue that generalizing such adversarial joint distribution matching to \emph{multiple} distributions and arbitrary number of random variables can unlock a much larger potential for representation learning and generative modeling that has not yet been explored by previous approaches~\cite{ali,alice,bigan}. Unlike other generative models such as VAEs, as we show, GANs can be generalized to match more than two joint distributions of tuples of an arbitrary number of random variables. We demonstrate that this allows integration of self-supervised learning and learned or prior knowledge about the properties of desired solutions. 

%For example, as argued in~\cite{alice}, one such known property for encoder-decoder based models is that of cycle-consistency~\cite{cyclegan}, i.e., the fact that with an ideal solution, the reconstruction of any image from the data distribution should be similar to the original image. %\piyush{I feel that we should not talk about ALICE or pixel-wise reconstruction here but in related work. Here we should just say what we are doing. The intro has gotten quite long} \YD{Sure sir. I've moved it to related work. Is the present length of the intro okay?}  %
Unlike previous approaches relying on pixel-level reconstruction objectives  \citep{vae,alice,age} which are known to be one of the causes of blurriness~\cite{alice,vae_understanding}, we propose an approach for incorporating multiple layers of reconstructive feedback within the adversarial joint distribution matching framework.  
%For example, using $L_2$ loss in VAEs is equivalent to assuming an isotropic Gaussian distribution in the pixel space conditioned on the latent variable, and minimization of this loss leads to averaging across different modes in the actual pixel space .%
%For example, in VAEs, using squared error loss is equivalent to assuming an isotropic Gaussian distribution in the pixel space conditioned on the latent variable. Minimization of this squared error is known to be one of the causes of blurriness in the generated images as the model ends up averaging across different modes in the pixel space \cite{multi}. 
%The notion of cycle consistency in adversarially learned inference~\cite{alice,cyclegan} can be generalized in various ways. For applications, such as inpainting and translation, it is particularly important that parts of original images be compatible or consistent with other parts of reconstructed images. %Unlike ALICE \piyush{[haven't talked about ALICE yet - maybe remove "Unlike ALICE"?]},% 
Our approach allows incorporation of such task-specific self-supervised feedback and knowledge of equivariance of reconstructions to dataset-dependent transformations within the framework of adversarial joint distribution matching, without requiring careful tuning of weighing terms for different objectives. In particular, we consider a discriminator that classifies joint samples of an arbitrary number of random variables into an arbitrary number of classes. Each class essentially represents a distribution over tuples of image and latent vectors, defined by recursively computing the encodings and their reconstructions. This provides multiple layers of information for each real image or sampled vector while allowing the generator-encoder pair to gain \emph{explicit} feedback on the quality and relevance of different types of reconstructions. Fig.~\ref{fig:main} illustrates this process through a diagram. In the rest of the paper, we refer to our proposed framework as \textbf{Generalized Adversarially Learned Inference (GALI)}.

  While Adversarially Learned Inference (ALI) can be generalized to multi-class classification within the framework of minimax likelihood based objective, the resulting training procedure is still susceptible to vanishing gradients for the generator-encoder. We illustrate this problem and devise an alternative objective that extends the non saturating GAN objective to multiple distributions. We develop a generalized framework for distribution matching with the following main contributions:
 % \begin{enumerate}[
%     topsep=0pt,
%     noitemsep,
%     leftmargin=*,
%     itemindent=12pt]
 \begin{enumerate}
    \item We introduce a scalable approach for incorporating multiple layers of knowledge-based, reconstructive and self-supervised feedback in adversarially learned inference without relying on fixed pixel or feature level similarity metrics.
    \item %We illustrate how a multi-class classification minimax objective suffers from the vanishing gradient problem, and to remedy this% 
    % \abhi{what do we mean by stability of the objective here? it will be good to clarify}\YD{the major issue with the original multi-class classification minimax objective is that it suffers from vanishing gradient for the generator when the discriminator's accuracy is high. "stable" is probably not the right word for it.}
    We propose a non-saturating objective 
    for training a generator network when the corresponding discriminator performs multi-class classification. We further prove that our proposed objective has the same global optima as the mini-max objective which matches all the distributions simultaneously. %which suggests its applicability for domain transformation and image translation tasks.%
    \item We  demonstrate how the proposed approach can incorporate pre-trained models and can naturally be adapted for particular tasks such as image inpainting by incorporating suitably designed auxiliary tasks within the framework of adversarial joint distribution matching.
\end{enumerate}
\section{Preliminaries}
\label{sec:prelim}
The following minimax objective  serves as the basic framework for optimization in the ALI/BiGAN framework.
\begin{equation}
\min\limits_{G, E}
\max\limits_D
V(D, E, G)
\label{eq:ali}
\end{equation}
where
\begin{align*}
V(D, E, G)
&\defeq
\mathbb{E}_{\x \sim p_\X}\big[
  \underbrace{
    \expect{\z \sim p_E(\cdot | \x)}{
      \log D(\x, \z)
    }
  }_{
    \log D(\x, E(\x))
  }
\big] \\&+
\mathbb{E}_{\z \sim p_\Z}\big[
  \underbrace{
    \expect{\x \sim p_G(\cdot | \z)}{
      \logp{1 - D(\x, \z)}
    }
  }_{
    \logp{1 - D(G(\z), \z)}
  }
\big].
\label{eq:vdge1}
\end{align*}
Here, the generator $G$ and encoder $E$ can either be both deterministic, such that $G:\OZ \to \OX$ with $p_G(\x | \z) = \delta\left(\x - G(\z)\right)$ and  $E:\OX \to \OZ$ with $p_E(\z | \x) = \delta\left(\z - E(\x)\right)$ or the encoder can be stochastic. Deterministic models were used in the BiGAN~\cite{bigan} approach, while a stochastic encoder was used in ALI~\cite{ali}. For all our experiments and discussions, we use a stochastic encoder following ALI~\cite{ali} but denote samples from $p_E(\z | \x)$ as $E(\x)$ for notational convenience.
Under the assumption of an optimal discriminator, minimization of the generator-encoder pair's objective is equivalent to minimization of the Jensen-Shannon (JS) divergence~\cite{JSD} between the two joint distributions. Thus, achieving the global minimum of the objective is equivalent to the two joint distributions becoming equal.

\section{The Proposed Approach}
\label{sec:proposed}
The proposed approach is based on the two sequences in Fig.~\ref{fig:main} : the top sequence starts from a real latent variable and its corresponding generation and contains all the subsequent reconstructions and their encodings while the bottom sequence starts from a real image and contains its corresponding set of reconstructions and encodings. Ideally, we wish all the latent vectors and all the images within a sequence to be identical. We argue that the optimization objective presented in the previous section is the simplest case of a general family of objectives where the discriminator tries to classify $n$ classes of tuples  of size $m$,
% (\HB{What are n and m?}) (\YD{n is the number of tuples and m is the size of each tuple. 
% For the example used for proofs and experiments, n = 4 and m =2})
of images and latent vectors with the variables within a tuple all belonging to one of the sequences in Fig~\ref{fig:main} while the generator tries to fool it to incorrectly classify.
By including additional latent vectors and images, we allow the generator-encoder pair to receive multiple layers of reconstructive feedback on each latent vector and each generated image while modifying the discriminator to an $n$-way classifier encourages it to perform increasingly fine grained discrimination between different image-latent variable tuples.  We experimentally demonstrate results for $(n=4,m=2)$,$(n=8,m=2)$, and $(n=8,m=4)$.
Our goal is to design an objective where the discriminator is tasked with discriminating against each of the joint distributions specified by the tuples, and the generator and encoder try to modify the joint distributions such that distributions of all the classes of tuples are indistinguishable from each other. We analyse different alternatives for the same in the subsequent sections.
%\HB{How to decide till when to continue the chain?}\YD{The computational requirements for training increase as the number of tuples and size of tuples increases. We experimentally demonstrate results for n=4,m=2, n=8,m=2 and n=8,m=4}
\subsection{Multiclass Classifier Discriminator}
\label{sec:multiclass}
We first consider the expected log-likelihood based mini-max objective for the case of $n=4$ classes and tuples of size $m=2$.
We choose the set of pairs (classes) to be: $(\x, E(\x)),(G(\z), \z),(\x, E(G(E(\x)))),(G(E(G(\z))), \z)$.
The discriminator is modified to perform multi-class classification with the output probabilities of input (image, latent vector) $(\x_{in},\z_{in})$ for the $i^{th}$ class denoted by $D_i(\x_{in},\z_{in})$.  So, the output of $D(\x_{in},\z_{in})$ is a vector $[D_1(\x_{in},\z_{in}),...D_i(\x_{in},\z_{in}),..]$ where $D_i(\x_{in},\z_{in})$ denotes the output probability for the $i^{th}$ class.
% \HB{Just to clarify, the pairs represent the two distributions, the samples from which the discriminator is supposed to discriminate against right? It will perhaps be good to clarify that - we want E(x), z, E(G(E(x))) to be indistinguishable? Similarly, we want x, G(z), G(E(G(z))) to be indistinguishable?}. \YD{Yeah but is also important to match the correspondence between x, E(x) and z, G(z). Hence, we are considering their joint distributions}
   %Following \citep{ali,alice} we consider architectures where the discriminator $D(\cdot)$ takes as input the $X$ and $z$ of each pair and processes them independently through two encoder networks, joins the concatenated encodings and passes that through a fully connected layers. The detailed architecture is describe in the Appendix.%
% \HB{Need to mention that x and z are processed independently by two networks, whose outputs are concatenated together and processed jointly for a few more layers --> i.e. mention how D processes the inputs}.  
The minimax objective with a multi-class classifier discriminator, following a straightforward generalization of ALI in Eq.~(\ref{eq:ali}) thus becomes:
\begin{equation}
\min\limits_{G, E}
\max\limits_D
V(D, E, G)
\label{eq:main}
\end{equation}
where
\begin{align}
\begin{split}
&V(D, E, G)
\defeq \mathbb{E}_{\x \sim p_\X}\big[
    \logp{D_1(\x, E(\x))}
\big] \\&+
\mathbb{E}_{\z \sim p_\Z}\big[
  {
    \logp{D_2(G(\z), \z)}
  }
\big]\\&
+
\mathbb{E}_{\x \sim p_\X}\big[
    \logp {D_3(\x, E(G(E(\x))))}
\big]\\&+ \mathbb{E}_{\z \sim p_\Z}\big[
    \logp {D_4(G(E(G(\z))), \z)}
\big].
\end{split}
\label{eq:vdge2}
\end{align}
Although the above adversarial game captures the multiple layers of reconstructive feedback described in Fig.~\ref{fig:main}, it is insufficient for stable training due to \textit{vanishing gradients}. Consider the gradients for the parameters of the generator  and the encoder with the above objective. Since the gradient of the softmax activation function w.r.t the logits vanishes whenever one of the logits dominates the rest, when the discriminator is able to classify accurately, the gradients of the generator-encoder's objective nearly vanish and the generator-encoder pair does not receive any feedback. In order to remedy this, we provide an alternate training objective for the generator-encoder based on products of the likelihoods. We first describe how the vanishing gradients problem cannot be avoided by using an objective based on misclassification likelihood.

\subsubsection{Misclassification Likelihood}
At first it might seem that a natural way to alleviate the vanishing gradient problem is to replace each log likelihood term $\logp{D_i(\x_{in},\z_{in})}$ in the mini-max generator-encoder minimization objective with the corresponding misclassification log-likelihood $\logp{1-D_i(\x_{in},\z_{in})}$ for the given class to construct a maximization objective. This is the approach used while designing the non-saturating objectives for standard GANs~\cite{nonsat1}, ALI, and BiGAN frameworks \citep{ali,bigan}.
However, in the multiclass classification framework, the value and the corresponding gradients for the misclassification objective can vanish even when the discriminator learns to accurately reject many of the incorrect classes as long as it has a low output probability for the true class. For example, for the 4 classes considered above, the discriminator may learn to accurately reject classes 3 and 4 for a pair belonging to class 1 but might still have a high misclassification likelihood if it incorrectly identifies the pair as belonging to class 2. As such an objective does not optimize the individual probabilities for the incorrect classes, the gradient for the generator-encoder pair would provide no feedback for causing an increase in the output probabilities for the remaining classes (3 and 4).

\subsubsection{Product of Terms}
\label{sec:product}
In light of the above, we desire to obtain an objective such that instead of just encouraging the generator-encoder to cause lower discriminator output probability for the \textit{right} class, the generator-encoder's gradient for the modified objective enforces each of the \textit{wrong} classes to have high output probabilities. With this goal, we propose the product of terms objective that explicitly encourages all the distributions to match. The proposed objective for the 4 classes considered above is given below: 
 \begin{equation}
\begin{split}
&\max\limits_{G, E}[
\mathbb{E}_{\x \sim p_\X}[
    \logp{D_2(\x, E(\x))D_3(\x, E(\x))D_4(\x, E(\x))}
]
\\
&+
\mathbb{E}_{\z \sim p_\Z}[
    \logp{D_1(G(\z), \z)D_3(G(\z), \z)D_4(G(\z), \z)}]
\\
&+
\mathbb{E}_{\x \sim p_\X}[
    \logp{D_1(\x, E(G(E(\x))))D_2(\x, E(G(E(\x))))}\\
&D_4(\x, E(G(E(\x))))
]
\\
&+
\mathbb{E}_{\z \sim p_\Z}[
    \logp{D_1(G(E(G(\z))), \z)D_2(G(E(G(\z))),\z)}\\
&D_3(G(E(G(\z))), \z)
]]
\end{split}
\label{eq:product}
\end{equation}
%\vskip -0.11
Since the vanishing gradients problem only arises in the generator-encoder, we use the same objective as \ref{eq:vdge2} for the discriminator. In Eq.~\ref{eq:product}, each of the terms of the form $\logp{D_i(\x_{in},\z_{in})D_j(\x_{in},\z_{in})D_k(\x_{in},\z_{in})}$ can be further split as $\logp{D_i(\x_{in},\z_{in})}+ \logp{D_j(\x_{in},\z_{in})} +\logp{D_k(\x_{in},\z_{in})}$. The above objective encourages the parameters of the generator and encoder to ensure that none of the incorrect classes are easily rejected by the discriminator.  %\HB{accurately? what do you mean}%
This is because the generator-encoder pair is explicitly trained to cause an increase in the discriminator output probabilities for all the \textit{wrong} classes.  
Moreover, the objective does not lead to vanishing gradients as discarding any of these classes as being true incurs a large penalty in terms of the objective and its gradient. %\HB{did not get this. why do we want large probability for the wrong classes? We may have to re-write the explanations a bit. Seems a bit hairy} %
In the next subsection and the appendix we show that the Product of Terms objective has a global optimum which matches all the joint distributions corresponding to the different classes of tuples simultaneously. We further demonstrate the effectiveness of this objective through experiments.
\subsection{The optimal discriminator}
The optimal discriminator $D^*$ for the discriminator's objective in Eq.~(\ref{eq:vdge2}), can be described as $D^*\defeq
\argmax_D V(D, E, G)
$, for any $E$ and $G$. Following the derivation in the appendix, we obtain the following functional form of $D^*$:
$
    D^*_{i}(\x_{in},\z_{in}) = \frac{p_i(\x_{in},\z_{in})}{\sum_{j=1}^4 p_j(\x_{in},\z_{in})}
$.

Here, $D_i$ corresponds to the output probability of the $i^{th}$ class among the four classes: $(\x, E(\x)),(G(\z), \z),(\x, E(G(E(\x)))),(G(E(G(\z))), \z)$. $p_i$ is the joint probability density of the corresponding $\x_{in}$ and $\z_{in}$ in each of the $(\x_{in},\z_{in})$ pairs above.
\subsection{The optimal Generator-Encoder for the Product of Terms Objective}
%Similar to the analysis for the minimzx, to obtain the global optima for the generator $G$ and encoder $E$, we substitute the optimal discriminator in each of the terms of the optimization objective .%
\Abhi{General comment: it's probably better to number the appendices as A,B,C.. and provide the appendix number wherever we reference the appendix so that it's easy to find for reviewers.}
Following the derivation provided in the appendix and  substituting the optimal discriminator found above  ($D^* = \argmax_D V(D,E,G)$) in the Product of Terms objective (Eq.~\ref{eq:product}) for the generator-encoder leads to the maximization objective:
$
C(G,E) \leq- \logp{4^9} - JSD_{\frac{1}{4},\frac{1}{4},\frac{1}{4},\frac{1}{4}}(p_1,p_2,p_3,p_4)
$.
where $JSD_{\frac{1}{4},\frac{1}{4},\frac{1}{4},\frac{1}{4}}$ denotes the generalized Jensen–Shannon divergence  ~\cite{JSD} with equal weights assigned to each distribution as described in the appendix.
Since the JSD for the four distributions is non negative and vanishes if and only if $p_1 = p_2 = p_3 = p_4$, the global optimum for the product of terms objective is given by:
$
    p_{(\x, E(\x))} = p_{(G(\z), \z)} = p_{(\x, E(G(E(\x))))} = p_{(G(E(G(\z))), \z)}
$
 This is the same as the optima of the original generalized objective in Eq.~(\ref{eq:main}) (appendix). However, our proposed objective matches all the distributions simultaneously without suffering from vanishing gradients.
%\abhi{why is distributing evenly among classes better?} \YD{Sorry, the emphasis is not on being evenly distributed but on no class being easily discarded as being true without a huge penalty.}%
\vspace*{-0.1cm}

\subsection{Extension to Arbitrary Number and Size of Tuples}
\label{sec:extra}
\vspace*{-0.05cm}
The analysis presented above can be extended to accommodate any number $n$ of tuples of arbitrary size $m$ from the two chains in \ref{fig:main}. We demonstrate this for the case of $n=8$ and $m=4$ for the SVHN dataset. We start with one tuple from each chain: $(\x, E(\x),G(E(\x)),E(G(E(\x))))$ and $(G(\z), \z,G(E(G(\z))),E(G(\z)))$ and construct additional tuples by permuting within the images and latent vectors for both of these tuples to give a total of $4\times2$ tuples.  %\abhi{we should discuss the advantages of working with these m-tuples}% 
These classes of 4-tuples allow the discriminator to directly discriminate between an image and its reconstruction in both image and latent space. The number of pairs $n$ to be considered and the size of tuples is limited only by computational cost, although intuitively we expect to see diminishing returns in terms of performance when increasing $n$ beyond a point as matching the distributions for two random variables enforces the matching of the subsequent chains.% For example, when the distribution of the reconstructed images closely matches that of the original images, the chain of corresponding distributions starting from $x$ and $G(E(x))$ in Fig.~\ref{fig:mainfig} are very similar such that discriminating between them provides little additional information for training. 
%
% \vspace*{-0.1cm}
% \begin{wrapfigure}[10]{r}{0.45\textwidth}
% \vspace*{-0.2cm}
% %\framebox[4.0in]{$\;$}
% \includegraphics[width=\linewidth]{mixed.png}
% \caption{Sections from reconstructions of real images corresponding to missing patches are combined with the original real images to form a set of \textit{mixed} images.}
% \label{fig:mixedmixed}
% \end{wrapfigure}
\vspace*{-0.1cm}
\begin{figure}[h!]
\vspace*{-0.2cm}
%\framebox[4.0in]{$\;$}
\includegraphics[width=\linewidth]{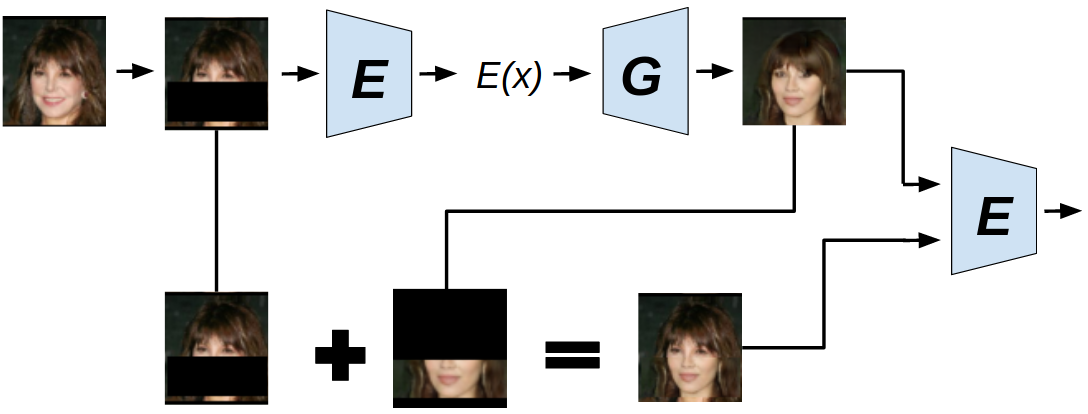}
\caption{Sections from reconstructions of real images corresponding to missing patches are combined with the original real images to form a set of \textit{mixed} images.}
\label{fig:mixedmixed}
\vspace*{-0.2cm}
\end{figure}

\subsection{Self-Supervised Feedback using Mixed Images}
\label{sec:mixed}
\vspace*{-0.05cm}
For tasks such as image inpainting and translation, an important desideratum for a reconstructed image is its consistency with parts of original image.  We demonstrate that such task specific feedback can be incorporated into the GALI framework through self-supervised learning tasks. We experiment with incorporating one such task into the reconstructive chain of Fig. ~\ref{fig:main}, namely ensuring consistency and quality of \textit{mixed} images constructed by combining the inpainted sections of reconstructions of real images with the other parts of the corresponding original real images. During training, we mask randomly sampled regions of the images input to the encoder, as shown in Fig.~\ref{fig:mixedmixed}.
Our analysis of the optimal solutions proves that the multi-class classifier based approach can be used to match any number of distributions of (image, latent variable) pairs, as long as each such pair is generated using the real data distribution, fixed prior for the noise or any operation on these such that the output's distribution depends only on generator-encoder parameters. Since the set of mixed images and their encodings satisfy these constraints, introducing them into the set of (image, latent vector) pairs used in the objective leads to matching of the distribution of mixed images with the real image distribution as well as the matching of correspondence between the encodings and reconstructions.
Thus, this type of feedback introduces an additional constraint on the model to ensure the consistency of the inpainted patch with the original image.  We show this empirically in the experiments section. 
The self-supervision based model is constructed by introducing 4 additional classes of (image,latent-vector) pairs into the GALI-4 model so that the discriminator is trained to perform an 8-way classification. These additional classes depend on a distribution of masks that are applied to the images. For our experiments, we sample a mask $M$ for CelebA 64X64 images as follows: first, the height $w$ and the width $h$ of the mask are both independently drawn uniformly from the set ${1,\cdots,64}$. Subsequently, the $x$ (horizontal) and $y$ (vertical) indices (index origin = 0) of the bottom-left corner of the mask are drawn uniformly from the set ${0,\cdots,63-h+1}$ and
${0,\cdots,63-w+1}$. The above procedure thus defines a probability distribution over masks $P(M)$ For every image input to the encoder, a new mask $M$ is sampled independently. Thus the four classes of distributions considered in GALI-4 are modified to classes whose samples are constructed as : $(\x, E(M_1(\x))),(G(\z), \z),(\x, E(M_2(G(E(M_1(\x))))))$,\ $(G((E(M_3(G(\z))), \z)$ where $M_1$, $M_2$, and $M_3$ are independently drawn from $P(M)$. For a sampled real image and its corresponding mask $M_1$, a mixed image $Mix(\x,M_1)$ is constructed by replacing the masked region of $\x$ by the corresponding region of $G(E(M_1(\x)))$. The additional 4 pairs then correspond to $(Mix(\x,M_1),E(M_1(\x))),(Mix(\x,M_1)$\ , $E(M_2(G(E(M_1(\x)))))),(\x, E(M_4(Mix(\x,M_1)))$ and $(G(E(M_1(\x)))),E(M_4(Mix(\x,M_1)))$ 
\subsection{Incorporation of Learned Knowledge}
\label{sec:knowledge}
Matching the real and generated data distributions should also result in the matching of the corresponding joint distributions of any of the extractable properties such as attributes, labels, perceptual features or segmentation masks. We argue that utilizing these outputs during training can impose additional constraints and provide additional information to the model similar to the reconstructive and self-supervised feedbacks discussed above. We demonstrate that our approach offers a principled way for incorporating these properties by introducing the final or hidden layer outputs of pretrained models as additional random variables in the tuples. For each class of tuples, these outputs could correspond to input any input image within the same chain as other images and latent variables in the tuple. 
For experiments and subsequent discussions, we consider the four classes of tuples from the objective in ~\ref{eq:product} and augment each with the respective outputs from a pre-trained model $M$ to obtain the tuple classes $(\x,E(\x),M(\x))$,$(G(\z), \z, M(G(\z)))$, $(\x, E(G(E(\x))),M(G(E(\x))),(G(E(G(\z))), \z, M(G(\z)))$. 
While feature level reconstructive feedback can be provided through $L_1$ or $L_2$ reconstruction objectives on the output features, our approach explicitly matches the joint distribution of these features with images and latent vectors of all classes (real, fake, reconstructed, etc.) Unlike, $L_1$ and $L_2$ based reconstruction terms which directly affect only the generator-encoder, our approach distributes the feedback from the pretrained model across all the components of the model including the discriminator. Moreover, as we demonstrate in the ablation study included in the appendix, by jointly providing reconstructive feedback on images, encodings and pretrained model activations, our model with learned knowledge outperforms the model based on  $L_2$ reconstruction of features on all metrics. Our approach can also be applied to any arbitrary type of model outputs such as segmentation masks, translations or labels without blurriness effects. %\HB{Can we state here that what we do is \textit{better} than L1,L2 recosnt objectives? can we say that L1/L2 losses would cause blurriness? Can we point to a section of experiments for this (not directly this, but related to this)}. Although we experiment with features of hidden layers, our approach can be applied to any arbitrary type of model outputs such as segmentation masks, translations and labels.%
%\HB{What?}%
\vspace*{-0.2cm}
\section{Related Work}
\vspace*{-0.05cm}
A number of VAE-GAN hybrids have been proposed in the recent years. Adversarial autoencoders \citep{aae} use GANs to match the aggregated posterior of the latent variables with the prior distribution instead of the KL-divergence minimization term in VAEs. VAE-GAN. \citep{vaegan} replaced the pixel-wise reconstruction error term with an error based on the features inferred by the discriminator. AVB \citep{avb} proposed using an auxiliary discriminator to approximately train a VAE model.  Adversarial Generator-Encoder Networks \citep{age} constructed an adversarial game directly between the generator and encoder without any discriminator to match the distributions in image and latent space. The model still relied on $L_2$ reconstruction to enforce cycle consistency. Unlike the above approaches, BIGAN\citep{bigan}, ALI\citep{ali} and our models do not rely on any reconstruction error term.
%ALICE \citep{alice} justified the use of reconstruction error terms in ALI as conditional entropy based regularization that constraints the form of latent variable-image joint distribution. We instead propose incorporating reconstructive feedback in a purely adversarial and generalizable manner.%
Although a recent work BigBiGAN~\cite{bigbigan} demonstrated how the ALI/BiGAN framework alone can allow achieving competitive representation learning, we emphasize that embedding \textit{more information} through multiple layers of feedback would further improve the inference. 
Some recent approaches~\cite{trianglegan,triplegan,jointgan} propose different frameworks for adversarially matching multiple joint distributions for domain transformation and conditional generation. We hypothesize that the use of our proposed product of terms objective and the proposed different types of feedback should lead to further improvements in these tasks similar to the improvements demonstrated for ALI in the experiments.

ALICE \citep{alice} illustrated how the ALI objective with stochasticity in both the generator and the encoder can easily result in an optimal solution where cycle consistency is violated ($x \neq \Tilde{x}$) due to the non-identifiability of solutions. The analysis however does not apply to our approach as our optimal solution explicitly matches additional joint distributions which involve reconstructions and their corresponding encodings. ALICE ~\citep{alice} further proposed to solve the above issue of non-identifiability  by adding conditional entropy regularization terms to the ALI objective. In practice, this regularization ends up being a pixel wise reconstruction objective such as the $L_2$ loss. Such objectives implicitly impose assumptions directly on the distribution of pixels conditioned on latent variables and are known to be one of the causes of blurriness in the generated images~\cite{vae_understanding,multi}.
% Limitations of the ALI/BIGAN objective were further highlighted by \citep{arora} who proved the existence of an optimal solution to the objective with meaningless encodings and finite support.
% \citep{hali} extended ALI to a hierarchical latent variable model. We leave the extension of our proposed approach with such hierarchical models to future work.

In augmented BiGAN~\citep{aug}, instead of  generalizing the discriminator to perform multiclass classification, the fake distribution is divided into two sources (one of generated images and latent vectors and the other of encodings of real images and their reconstructions) and a weighted average of the likelihood of these two parts is used for the dicriminator and generator's objective. Unlike our optimal solution which matches all the distributions simultaneously, the augmented BiGAN's optimal solution matches the (real image, encoding) distribution with the average of the two ``fake" distributions. This solution causes a trade-off between good reconstructions and good generation, which is avoided in our method since all the distributions are enforced to match simultaneously.
%While pretrained models are routinely used to improve performance in several supervised tasks such as classification, there use in generative modelling has largely been limited. Recent work \citep{perceptual} demonstrated the effectiveness of utilizing perceptual features from pre-trained models for training generative models. However, the proposed approach is limited to matching the first two moments of the features corresponding to real and generated data and relies on the universality of features of the pretrained models. %
%Due to the paucity of labeled data, self-supervised learning, and in particular self-supervised representation learning methods have become quite popular and necessary in recent times. In %
% \citep{NC} adversarially trained an autoencoder to generate any required part of an image given any subset of input pixels. However, the absence of an independent generator with a fixed latent space prior allows the distribution over reconstructed images to have a different support than the distribution over pure generations. Our approach incorporates such self-supervised feedback within the ALI framework and thus allows inference over a fixed latent space prior which is an important requirement for tasks such as interpolation.  
\vspace*{-0.3cm}
\section{Experiments}
\label{sec:exp}
\vspace*{-0.05cm}
Through experiments on two benchmark datasets, SVHN~\cite{svhn} and CelebA~\cite{celeba}, we aim to assess the reconstruction quality, meaningfulness of the learned representations for use in downstream tasks and generation, effects of extending the approach to more classes of tuples and larger tuple size, the ability of the proposed approach to incorporate knowledge from pretrained models trained for a different task, and its adaptability to specific tasks such as inpainting. We will make the experimental code publicly available.
\subsection{Notation and Setup}
 For all the experiments,following the description in the section ``The Proposed Approach", GALI-4 is the proposed GALI model with 4 terms, GALI-8 is the proposed GALI model with 8 terms and GALI-PT is the GALI-4 model augmented with a pretrained network $M$.

For all our proposed models and both ALI~\cite{ali} and ALICE~\cite{alice} (ALI + $L_2$ reconstruction error) baselines, we borrow the architectures from \citep{ali} with the discriminator using spectral normalization~\cite{spectral} instead of batch normalization and dropout~\cite{dropout}. As shown in Table 3 and Figure 3, our baseline obtains similar representation learning scores and reconstruction quality as reported in ALI \citep{ali}. All the architectural details and hyper-parameters considered are further described in the appendix.
\vspace*{-0.3cm}
\subsection{Reconstruction Quality}
We evaluate the reconstruction quality on test images using both pixel level and semantic similarity metrics. For pixel level similarity, we report the average pixel wise mean squared error on test datasets for SVHN and CelebA. For semantic similarity, we use the mean squared error of the features from pre-trained multi-digit classification model~\cite{multidig} trained on SVHN and a pre-trained attribute classification model for SVHN and CelebA datasets respectively. Further details of these models are provided in the appendix.

\begin{figure*}[h!]
\centering
 \begin{subfigure}[b]{0.40\textwidth}
               \centering
    \includegraphics[width=\textwidth]{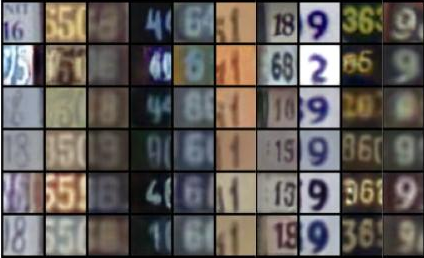}
    \caption{SVHN dataset.}
    \label{fig:rec_svhn}
        \end{subfigure}\hspace*{0.1cm}
                 \begin{subfigure}[b]{0.49\textwidth}
               \centering
    \includegraphics[width=\textwidth]{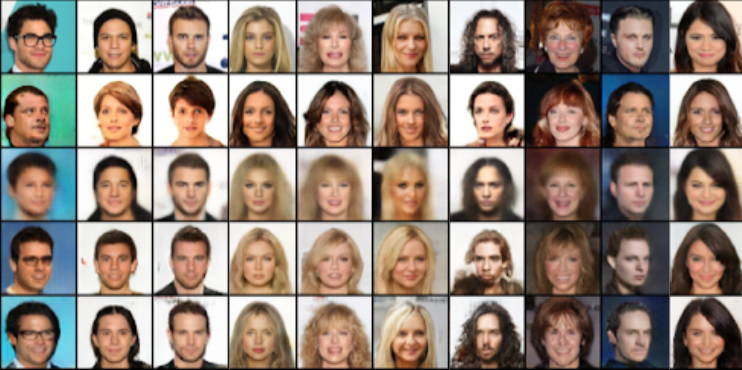}
    \caption{CelebA dataset.}
    \label{fig:rec_celeba}
        \end{subfigure}

            %\quad
     \caption{\textit{Top to bottom}: Original images, images reconstructed by ALI, ALICE, GALI-4, GALI-8, and GALI-PT for (a) and ALI, ALICE, GALI-4, and GALI-PT for (b). }  
 \label{fig:rec_all}
 \vspace*{-0.2cm}
\end{figure*}

\begin{table}[h!]
%  \begin{minipage}[c]{0.27\textwidth}
% \caption{\small{Pixel-wise Mean-Squared Error (MSE) and Feature level Mean Squared Error (MSE) on the test set for SVHN. Lower is better.}}
% \label{sample-table}
% \end{minipage}
%\vskip 0.05in
\caption{\small Pixel-wise Mean-Squared Error (MSE) and Feature level Mean Squared Error (MSE) on the test set for SVHN (left table) and CelebA (right table). Lower is better. \Abhi{this table doesn't look pretty. If there is time, consider reformatting to combine two tables into one.. maybe something like table 1 in https://arxiv.org/pdf/2002.08927.pdf. Also, table caption should be above the table according to NIPS formatting guidelines.}}
\label{tb:reconsttwo}
 \begin{minipage}[c]{0.52\textwidth}
\begin{center}
\begin{small}
\begin{sc}
\begin{tabular}{lcccr}
\toprule
Model  & pixel-MSE & feature-MSE\\
\midrule
ALI &  $0.023 \pm 0.0024$ & $0.0857 \pm0.0075$\\
ALICE & $0.0096 \pm 0.00067$ & $0.0736\pm 0.0058 $\\
GALI-4 & $0.0132 \pm 0.0011$  &$0.0717 \pm 0.0067$\\
GALI-8 & $\mathbf{0.0095} \pm 0.0016$ & $\mathbf{0.066} \pm 0.0060$ \\
GALI-pt & $\mathbf{0.0093} \pm 0.00099$ & $ \mathbf{0.041} \pm 0.0046$\\
\bottomrule
\end{tabular}
\end{sc}
\end{small}
\end{center}
% \vskip -0.1in
% \label{tb:reconst1}
\end{minipage}
%\vspace*{-0.2cm}
% \end{table}
% \begin{table}[htb]
%  \begin{minipage}[c]{0.27\textwidth}
% \caption{\small Pixel-wise Mean-Squared Error (MSE) and Feature level Mean Squared Error (MSE) on the test set for CelebA. Lower is better.}
% \label{sample-table}
% \end{minipage}
% %\vskip 0.15in
 \begin{minipage}[c]{0.52\textwidth}
\begin{center}
\begin{small}
\begin{sc}
\begin{tabular}{lcccr}
\toprule
Model  & pixel-MSE & feature-MSE\\
\midrule
ALI &  $0.074 \pm 0.004$  & $0.307  \pm 0.018$ \\
ALICE &  $0.042 \pm 0.002$& $0.248 \pm
0.013$\\
GALI-4 & $\mathbf{0.036} \pm 0.002$ & $\mathbf{0.201} \pm 0.012$ \\
GALI-pt &$\mathbf{0.032} \pm 0.0019$ & $\mathbf{0.131} \pm 0.008$\\
\bottomrule
\end{tabular}
\end{sc}
\end{small}
\end{center}
% \vskip -0.1in
\end{minipage}
\vspace*{-0.3cm}
\end{table}

\begin{table}
\setlength{\tabcolsep}{0.5pt}
\vspace{-1em}
\caption{Missclassification rate of GALI-4 and various baselines on the test set of SVHN dataset demonstrating the usefulness of learned representations. The results for baselines are from ~\cite{ali}. Lower is better.\vspace*{-0.4cm}}

\label{sample-table}

\begin{center}
\begin{small}
\begin{tabular}{lcccr}
\toprule
Model  & Misc rate (\%) \\
\midrule
VAE~\cite{vae} & 36.02\\
%SWWAE with dropout~\cite{zhao} & 23.56\\
DCGAN + L2-SVM & 22.18\\
SDGM~\cite{sdgm} & 16.61 \\
NC ~\cite{NC} & 17.12\\
ALI \citep{ali} & 19.11 $\pm$ 0.50 \\
ALI (baseline) & 19.05 $\pm$ 0.53 \\
%ALI ~\cite{ali} & 19.43 $\pm$ 0.51 \\
ALICE ~\cite{alice} & 18.89 $\pm$ 0.52\\ 
GALI-4  & \textbf{16.58 $\pm$ 0.38}\\
GALI-8 & \textbf{15.82 $\pm$ 0.43}\\
GALI-PT (supervised) & \textbf{11.43 $\pm$ 0.30}\\
\bottomrule
\end{tabular}
%\end{small}
\end{small}
\end{center}
\vskip -0.1in
\label{tb:representations}
%\vspace*{0.5cm}
\end{table}

% \begin{wraptable}[18]{r}{0.4\textwidth}
% \setlength{\tabcolsep}{3pt}
% \vspace{-1em}
% \caption{Missclassification rate of GALI-4 and various baselines on the test set of SVHN dataset demonstrating the usefulness of learned representations. The results for baselines are from ~\cite{ali}. Lower is better.\vspace*{-0.4cm}}

% \label{sample-table}

% \begin{center}
% \begin{small}
% \begin{tabular}{lcccr}
% \toprule
% Model  & Misc rate (\%) \\
% \midrule
% VAE~\cite{vae} & 36.02\\
% %SWWAE with dropout~\cite{zhao} & 23.56\\
% DCGAN + L2-SVM~\cite{dcgan} & 22.18\\
% SDGM~\cite{sdgm} & 16.61 \\
% NC ~\cite{NC} & 17.12\\
% ALI \citep{ali} & 19.11 $\pm$ 0.50 \\
% ALI (baseline) & 19.05 $\pm$ 0.53 \\
% %ALI ~\cite{ali} & 19.43 $\pm$ 0.51 \\
% ALICE ~\cite{alice} & 18.89 $\pm$ 0.52\\ 
% GALI-4  & \textbf{16.58 $\pm$ 0.38}\\
% GALI-8 & \textbf{15.82 $\pm$ 0.43}\\
% GALI-PT (supervised) & \textbf{11.43 $\pm$ 0.30}\\
% \bottomrule
% \end{tabular}
% %\end{small}
% \end{small}
% \end{center}
% \vskip -0.1in
% \label{tb:representations}
% %\vspace*{0.5cm}
% \end{wraptable}

The test set reconstructions on SHVN and CelebA $(64\times64)$ datasets for the proposed models, the ALI baseline, and the ALICE baseline are shown in Table~\ref{tb:reconsttwo}.
As reported in \citep{alice}, the improvements in reconstructions for ALICE are obtained at the cost of blurriness in images. The blurriness is visible in the reconstructions shown in Fig.~\ref{fig:rec_svhn} and Fig.~\ref{fig:rec_celeba} and quantitatively verified through the higher FID score of ALICE in Table~\ref{tb:fid} as well as the higher values of the feature level MSE for both datasets. Moreover, we observe that the introduction of L2 reconstruction error does not lead to significant gains in the usefulness of the learned representations for downstream tasks. Our models achieve significant improvements in the quality of reconstructions and representations over the ALI baseline without succumbing to the above drawbacks.
\vspace*{-0.2cm}
\subsection{Representation Learning}
In Table~\ref{tb:representations}, we evaluate the representation learning capabilities of the encoder by training a linear SVM model on features corresponding to 1000 labelled images from the training set. Following \citep{ali}, the feature vectors are obtained by concatenating the last three hidden layers of the encoder as well as its output.The hyperparameters of the SVM model are selected using a held-out validation set. We report the average test set misclassification rate for 100 different SVMs trained on different random 1000-example training sets. We also report the misclassification rate for other semi-supervised learning approaches for comparison. 
\vspace*{-0.3cm}
% \begin{table}[htb]
% \setlength{\tabcolsep}{3pt}
% \begin{minipage}[c]{0.27\textwidth}
% \caption{Missclassification rate of GALI-4 and various baselines on the test set of SVHN dataset demonstrating the usefulness of learned representations. The results for baselines are obtained from ~\cite{ali}. Lower is better.}
% \end{minipage}
% \label{sample-table}
% \begin{minipage}[c]{0.57\textwidth}
% \begin{center}
% \begin{small}
% \begin{tabular}{lcccr}
% \toprule
% Model  & Misc rate (\%) \\
% \midrule
% VAE~\cite{vae} & 36.02\\
% %SWWAE with dropout~\cite{zhao} & 23.56\\
% DCGAN + L2-SVM~\cite{dcgan} & 22.18\\
% SDGM~\cite{sdgm} & 16.61 \\
% NC ~\cite{NC} & 17.12\\
% ALI \citep{ali} & 19.11 $\pm$ 0.50 \\
% ALI (baseline) & 19.05 $\pm$ 0.53 \\
% %ALI ~\cite{ali} & 19.43 $\pm$ 0.51 \\
% ALICE ~\cite{alice} & 18.89 $\pm$ 0.52\\ 
% GALI-4  & \textbf{16.58 $\pm$ 0.38}\\
% GALI-8 & \textbf{15.82 $\pm$ 0.43}\\
% GALI-pretrain (supervised) & \textbf{11.43 $\pm$ 0.30}\\
% \bottomrule
% \end{tabular}
% %\end{small}
% \end{small}
% \end{center}
% \vskip -0.1in
% \label{tb:representations}
% \end{minipage}
% \vspace*{0.1cm}
% \end{table}

\subsection{Image Generation Quality}
It is important to ensure that improved reconstructions do not come at the cost of poor generation quality. Such a trade-off is possible when the encoder-generator pair learns to encode pixel level information. It is instead desired that variations in the latent space of the generator correspond to variations in explanatory factors of variation of the data. We evaluate our model's image generation ability using the Frechet Inception Distance (FID) metric \cite{fid} on the CelebA dataset in Table~\ref{tb:fid}. We provide visualizations of the generated samples in the appendix.

%\vspace*{-0.2cm}
\subsection{Image Inpainting}
\begin{figure}[h]
\begin{minipage}[c]{0.43\textwidth}
\begin{center}
%\framebox[4.0in]{$\;$}
\includegraphics[width=0.87\linewidth]{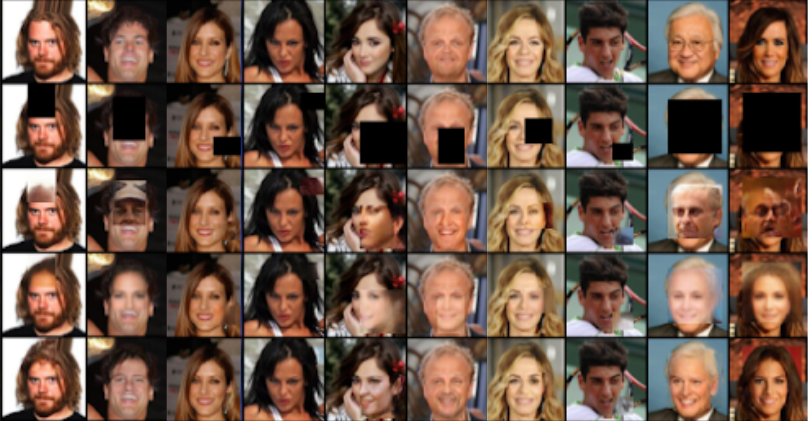}
\end{center}

\end{minipage}
\begin{minipage}[c]{0.43\textwidth}
%\begin{table}[htb]
% \caption{\small Pixel-wise Mean-Squared Error (MSE) and Feature level Mean Squared Error (MSE) on the image-inpainting task for CelebA. Lower is better.}
% \label{sample-table}
\vskip 0.15in
\begin{center}
\begin{small}
\begin{sc}
\begin{tabular}{lcccr}
\toprule
Model  & pixel-MSE & feature-MSE\\
\midrule
ALI &  $0.060 \pm
0.0080$  & $0.38 \pm
0.016$ \\
ALICE &  $0.047 \pm 0.0069$ &
$0.32 \pm
0.015$\\
GALI-4 & $0.046 \pm
0.0066$ &
$0.285 \pm
0.012$\\
GALI-mix & $\mathbf{0.031} \pm 0.0050$
& $\mathbf{0.164} \pm 0.010$\\
\bottomrule
\end{tabular}
\end{sc}
\end{small}
\end{center}
% \vskip -0.1in
% \vspace*{-0.3cm}
%\end{table}
\end{minipage}
\caption{\textbf{Left.} \textit{Top to bottom}: Original images, incomplete input images with the blackened region denoting an applied random occlusion mask, inpainted images from ALI , ALICE and GALI-4 with mixed images. \textbf{Right.} Pixel-wise Mean-Squared Error (MSE) and Feature level Mean Squared Error (MSE) on the image-inpainting task for CelebA. Lower is better.}
\label{fig:inpainting}
\vspace*{-0.5cm}
\end{figure}

%\vspace*{-0.2cm}
\label{sec:inpaintingexp}
In Fig.~\ref{fig:inpainting}, we evaluate our mixed images based model on an image inpainting task for the CelebA dataset through comparisons with ALI and ALICE baselines trained with the same procedure of masking out inputs. The ALICE based model leads to blurriness whereas ALI suffers from poor consistency with the original images. Our approach alleviates both these issues. Quantitative comparisons for this task are also described in Fig.~\ref{fig:inpainting}.

% %\vspace*{-0.4 cm}
% \begin{wrapfigure}{r}{0.7\linewidth}
% %\begin{figure}[h]
% \begin{center}
% %\framebox[4.0in]{$\;$}
% \includegraphics[width=\linewidth]{inpaint.png}
% \end{center}
% \caption{\textit{Top to bottom}: Original images, incomplete input images with the blackened region denoting an applied random occlusion mask, inpainted images from ALI , ALICE and GALI-4 with mixed images.}
% \label{fig:inpainting}
% %\end{figure}
% \end{wrapfigure}

Similar to the evaluation of reconstruction quality, we quantitatively evaluate image in painting capabilities of the models using both pixel and feature level metric. In the table in Fig.~\ref{fig:inpainting}, pixel-level MSE denotes the average pixel wise squared difference of the inpainted region with the original image, while the feature level MSE is calculated as the average squared difference between the feature vectors of the inpainted image and the original image for each model. The feature vectors are calculated using the same pretrained classifier as used for the reconstruction quality. In the table, we denote our proposed mixed images based model as GALI-mix while ALI and ALICE denote models obtained by utilizing the same distribution over masks as GALI-mix while feeding images to the encoder in the respective baselines. GALI-4 similarly denotes the GALI-4 model augmented with masked inputs. Results for GALI-4 without self-supervision are included as part of an ablation study in the appendix.% \HB{What does this last line mean?}%
% \begin{wraptable}{r}{0.6\l7inewidth}
% %\begin{table}[htb]
% \caption{\small Pixel-wise Mean-Squared Error (MSE) and Feature level Mean Squared Error (MSE) on the image-inpainting task for CelebA. Lower is better.}
% \label{sample-table}
% \vskip 0.15in
% \begin{center}
% \begin{small}
% \begin{sc}
% \begin{tabular}{lcccr}
% \toprule
% Model  & pixel-MSE & feature-MSE\\
% \midrule
% ALI &  $0.060 \pm
% 0.0080$  & $0.38 \pm
% 0.016$ \\
% ALICE &  $0.047 \pm 0.0069$ &
% $0.32 \pm
% 0.015$\\
% GALI-4 & $0.046 \pm
% 0.0066$ &
% $0.285 \pm
% 0.012$\\
% GALI-mix & $\mathbf{0.031} \pm 0.0050$
% & $\mathbf{0.164} \pm 0.010$\\
% \bottomrule
% \end{tabular}
% \end{sc}
% \end{small}
% \end{center}
% \vskip -0.1in
% \vspace*{-0.3cm}
% %\end{table}
% \end{wraptable}
% %\vspace*{-0.8cm}

% \begin{wraptable}[10]{r}{3.5cm}
% %\vspace*{-2cm}
% \caption{FID scores on CelebA demonstrating image generation quality of GALI-4, GALI-pretrain and the baselines. Lower is better.\vspace*{-0.2cm}}
% \label{sample-table}
% \begin{center}
% \begin{small}
% \begin{sc}
% \begin{tabular}{lcccr}
% \toprule
% Model  & FID\\
% \midrule
% ALI &  24.49 \\
% ALICE  & 36.91\\
% GALI-4  & \textbf{23.11}\\
% GALI-PT & \textbf{10.13}\\
% \bottomrule
% \end{tabular}
% \end{sc}
% \end{small}
% \end{center}
% \vskip -0.20in
% \label{tb:fid}
% \end{wraptable}

\begin{table}
%\vspace*{-2cm}
\begin{minipage}[c]{0.43\textwidth}
\caption{FID scores on CelebA demonstrating image generation quality of GALI-4, GALI-pretrain and the baselines. Lower is better.\vspace*{-0.2cm}}
\label{tb:fid}
\end{minipage}
\begin{minipage}[c]{0.43\textwidth}
\begin{center}
\begin{small}
\begin{sc}
\begin{tabular}{lcccr}
\toprule
Model  & FID\\
\midrule
ALI &  24.49 \\
ALICE  & 36.91\\
GALI-4  & \textbf{23.11}\\
GALI-PT & \textbf{10.13}\\
\bottomrule
\end{tabular}
\end{sc}
\end{small}
\end{center}
\vskip -0.20in
\end{minipage}
\end{table}

\vskip -0.20in
\vspace*{0.5cm}
\subsection{Utilization of Pretrained Models}
 For SVHN, the pretrained model M outputs features from the pre-trained multi-digit classification model used for the feature-level MSE above while for CelebA, M outputs the features from the pretrained inception net~\cite{inception_v3} used for calculating the FID scores. We emphasize that our goal is to demonstrate the ability of our approach to incorporate \textit{learned knowledge} from other models. These approaches however do not correspond to truly unsupervised settings as the pretrained models utilize supervision in the form of labelled SVHN digits and imagenet labels for SVHN and CelebA datasets respectively. This leads to expected yet significant improvements in metrics based on the output features of or the same training tasks as the pre-trained models such as misclassification rate and feature-MSE for SVHN and FID for CelebA. The above results demonstrate that the improvements further carry over to other independent metrics such as feature and pixel-MSE for CelebA and pixel-MSE for SVHN, indicating significant improvements in the  overall reconstruction and image generation quality.
\vspace*{-0.25cm}
\section{Conclusion}
\vspace*{-0.05cm}
In this paper, we proposed a novel framework for incorporating different types of generalized feedbacks in adversarially learned inference along with a non-saturating ``Product of Terms" objective for adversarially matching multiple distributions. Through experiments on two benchmark datasets, SVHN and CelebA, we demonstrated the efficacy of the proposed framework in terms of improvements in reconstruction quality, representation learning, image generation, and image inpainting as compared to previous approaches. 
% \bibliographystyle{plainnat}

%\newpage

\bibliography{example_paper}
\onecolumn

\section{Appendix}

\subsection{Analysis of Optimal Solutions}
In this section we derive closed form solutions for the optimal Discriminator and the optimal Generator-Encoder for the original mini-max and the product of terms objective. We first begin by providing a few clarifications:\\
$p_1,p_2,p_3,p_4$ are the joint distributions corresponding to the four classes of pairs of images and latent variables. Thus if 
$\Omega \defeq \OX \times \OZ$ is the joint latent and data space, then for a region $R \subseteq \Omega$, $p_i(R) \defeq \smallint_{\Omega} p_i(\x, \z) \ldeltafunc{(\x, \z) \in R} \d(\x, \z)$ for $i \in \{1,2,3,4\}$ are probability measures over that region. 
Since we use a stochastic encoder following ALI~\cite{ali} (even though we denote samples from $p_E(\z | \x)$ as $E(\x)$ for notational convenience), we use the reparametrization trick
\citep{vae} to backpropagate through the expectation over  $p_E(\z | \x)$ i.e if $p_E(\z | \x) = \mathcal{N}(\mu_E(x),
\sigma_E^2(x)I)$, we sample $z$ as $z = \mu_E(x) + \sigma_E(x) \odot \epsilon, \quad
    \epsilon \sim \mathcal{N}(0, I)$

In all the subsequent discussions, $x_{in},z_{in}$ are used to denote arbitrary images and latent vector pairs fed to the discriminator to distinguish them from $x$ and $z$ sampled from $p(x)$ and $p(z)$ respectively.

\subsubsection{Optimal Discriminator}
\label{eq:optdiscrim}
For any $E$ and $G$, the optimal discriminator $
D^*\defeq
\argmax_D V(D, E, G)
$ 
\begin{align}
    D^*_{i}(\x_{in},\z_{in}) = \frac{p_i(\x_{in},\z_{in})}{\sum_{j=1}^4 p_j(\x_{in},\z_{in})}
\end{align}

where $p_i$ corresponds to the distribution of the $i^{th}$ class among the four classes: $(\x, E(\x)),(G(\z), \z),(\x, E(G(E(\x)))),(G(E(G(\z))), \z)$. These distributions are fixed for fixed distributions $p(x)$ and $p(z)$, a fixed deterministic function $G(z)$ and a fixed function $E(x)$ defining the conditional distribution of the encoding given an input image.

\noindent \textbf{Proof:} For fixed $E$ and $G$, the maximization objective for $D$ is given by:
\begin{align*}
V(D,E,G) &= \int_{(\x_{in},\z_{in})} \big(p_1(\x_{in},\z_{in})\logp{D_1(\x_{in}, \z_{in})} + p_2(\x_{in},\z_{in})\logp{D_2(\x_{in},\z_{in})}
\\&+p_3(\x_{in},\z_{in})\logp {D_3(\x_{in},\z_{in})}  +p_4(\x_{in},\z_{in})\logp {D_4(\x_{in},\z_{in})}
\big)d(\x_{in},\z_{in})\\
\end{align*}
For a fixed $(\x_{in},\z_{in})$, the above integrand can be expressed in terms of the cross-entropy between two discrete distributions $p$ and $q$ with the support ${1,2,3,4}$ given by $p(i) = \frac{p_i(\x_{in},\z_{in})}{\sum_{j=1}^4 p_j(\x_{in},\z_{in})}$ and $q(i) =  D_i(\x_{in},\z_{in})$. We have:
\begin{align*}
    & p_1(\x_{in},\z_{in})\logp{D_1(\x_{in},\z_{in})(\x_{in},\z_{in})}+ p_2(\x_{in},\z_{in})\logp{D_2(\x_{in},\z_{in})} \\
&+p_3(\x_{in},\z_{in})\logp {D_3(\x_{in},\z_{in})}
+p_4(\x_{in},\z_{in})\logp {D_4(\x_{in},\z_{in})}\\
&= - (\sum_{j=1}^4 p_j(\x_{in},\z_{in}))\sum_{i=1}^4 p(i)\log q(i)\\
& = - (\sum_{j=1}^4 p_j(\x_{in},\z_{in}))H(p,q)\\
\end{align*}
Since $H(p,q) = H(p) + D_{KL}(p||q)$ is minimized w.r.t $q$ at $q=p$, we obtain the given formula for the optimal discriminator $D^*$. 
\subsubsection{Optimal Generator-Encoder for Minimax Objective}
Substituting the optimal discriminator for each $(G,E)$, the minimization objective for $(G,E)$ becomes:
\begin{align*}
    C(G,E) &= \max_D V(D,E,G)\\
    &= \mathbb{E}_{(\x_{in},\z_{in}) \sim p_1}\big[
    \logp{\frac{p_1(\x_{in},\z_{in})}{\sum_{j=1}^4 p_j(\x_{in},\z_{in})}}
\big]
+
\mathbb{E}_{(\x_{in},\z_{in}) \sim p_2}\big[
    \logp{\frac{p_2(\x_{in},\z_{in})}{\sum_{j=1}^4 p_j(\x_{in},\z_{in})}}
\big]
\\
&+
\mathbb{E}_{(\x_{in},\z_{in}) \sim p_3}\big[
    \logp{\frac{p_3(\x_{in},\z_{in})}{\sum_{j=1}^4 p_j(\x_{in},\z_{in})}}
\big]
+
\mathbb{E}_{(\x_{in},\z_{in}) \sim p_4}\big[
    \logp{\frac{p_4(\x_{in},\z_{in})}{\sum_{j=1}^4 p_j(\x_{in},\z_{in})}}
\big].\\
&= -\logp{256} + \mathbb{E}_{(\x_{in},\z_{in}) \sim p_1}\big[
    \logp{\frac{p_1(\x_{in},\z_{in})}{\frac{\sum_{j=1}^4 p_j(\x_{in},\z_{in})}{4}}}
\big]
+
\mathbb{E}_{(\x_{in},\z_{in}) \sim p_2}\big[
    \logp{\frac{p_2(\x_{in},\z_{in})}{\frac{\sum_{j=1}^4 p_j(\x_{in},\z_{in})}{4}}}
\big]
\\
&+
\mathbb{E}_{(\x_{in},\z_{in}) \sim p_3}\big[
    \logp{\frac{p_3(\x_{in},\z_{in})}{\frac{\sum_{j=1}^4 p_j(\x_{in},\z_{in})}{4}}}
\big]
+
\mathbb{E}_{(\x_{in},\z_{in}) \sim p_4}\big[
    \logp{\frac{p_4(\x_{in},\z_{in})}{\frac{\sum_{j=1}^4 p_j(\x_{in},\z_{in})}{4}}}
\big]\\
&= -\logp{256} + KL\left(p_1||\frac{\sum_{j=1}^4 p_j(\x_{in},\z_{in})}{4}\right) + KL\left(p_2||\frac{\sum_{j=1}^4 p_j(\x_{in},\z_{in})}{4}\right) \\
&+ KL\left(p_3||\frac{\sum_{j=1}^4 p_j(\x_{in},\z_{in})}{4}\right) + KL\left(p_1||\frac{\sum_{j=1}^4 p_j(\x_{in},\z_{in})}{4}\right)\\
&= -\logp{256} + JSD_{\frac{1}{4},\frac{1}{4},\frac{1}{4},\frac{1}{4}}(p_1,p_2,p_3,p_4)\\
\end{align*}
Since each of the KL terms and correspondingly the JSD above is non-negative, we have:
\begin{align*}
    C(G,E) \geq -\logp{256}
\end{align*}
Moreover, all the KL terms vanish simultaneously and correspondingly the JSD if and only if $p_1 = p_2 = p_3 = p_4$, leading to the global optimum given by:
\begin{align*}
    p_{(\x, E(\x))} = p_{(G(\z), \z)} = p_{(\x, E(G(E(\x))))} = p_{(G(E(G(\z))), \z)}
\end{align*}
Thus the minimax objective enforces the model to simultaneously match the above 4 distributions. 
\subsubsection{Optimal Generator-Encoder for Product of Terms Objective}
When the Generator-Encoder pair's is trained to maximize the modified Product of Terms objective, the optimal discriminator given a generator-encoder pair is still given by equation 3. Substituting the optimal discriminator in the maximization objective, we obtain:
\begin{align*}
    &C(G,E) =
\mathbb{E}_{(\x_{in},\z_{in}) \sim p_1}\big[
    \logp{\frac{p_2(\x_{in},\z_{in})p_3(\x_{in},\z_{in})p_4(\x_{in},\z_{in})}{(\sum_{j=1}^4 p_j(\x_{in},\z_{in}))^3}}\big]
\\&+
\mathbb{E}_{(\x_{in},\z_{in}) \sim p_2}\big[
    \logp{\frac{p_1(\x_{in},\z_{in})p_3(\x_{in},\z_{in})p_4(\x_{in},\z_{in})}{(\sum_{j=1}^4 p_j(\x_{in},\z_{in}))^3}}\big]
\\
&+
\mathbb{E}_{(\x_{in},\z_{in}) \sim p_3}\big[
    \logp{\frac{p_1(\x_{in},\z_{in})p_2(\x_{in},\z_{in})p_4(\x_{in},\z_{in})}{(\sum_{j=1}^4 p_j(\x_{in},\z_{in}))^3}}\big]
\\&+
\mathbb{E}_{(\x_{in},\z_{in}) \sim p_4}\big[
    \logp{\frac{p_1(\x_{in},\z_{in})p_2(\x_{in},\z_{in})p_3(\x_{in},\z_{in})}{(\sum_{j=1}^4 p_j(\x_{in},\z_{in}))^3}}\big]\\
&= -\logp{4^9}+\\ &\mathbb{E}_{(\x_{in},\z_{in}) \sim p_1}\big[
\logp{p_2(\x_{in},\z_{in})} + \logp{p_3(\x_{in},\z_{in})} + \logp{p_4(\x_{in},\z_{in})} -3\logp{\frac{\sum_{j=1}^4 p_j(\x_{in},\z_{in})}{4}}
\big]
+\\
&\mathbb{E}_{(\x_{in},\z_{in}) \sim p_2}\big[
    \logp{p_1(\x_{in},\z_{in})} + \logp{p_3(\x_{in},\z_{in})} + \logp{p_4(\x_{in},\z_{in})} -3\logp{\frac{\sum_{j=1}^4 p_j(\x_{in},\z_{in})}{4}}
\big]+\\
\end{align*}
\begin{align*}
&\mathbb{E}_{(\x_{in},\z_{in}) \sim p_3}\big[
     \logp{p_1(\x_{in},\z_{in})} + \logp{p_2(\x_{in},\z_{in})} + \logp{p_4(\x_{in},\z_{in})} -3\logp{\frac{\sum_{j=1}^4 p_j(\x_{in},\z_{in})}{4}}
\big]
+\\
&\mathbb{E}_{(\x_{in},\z_{in}) \sim p_4}\big[
    \logp{p_1(\x_{in},\z_{in})} + \logp{p_2(\x_{in},\z_{in})} + \logp{p_3(\x_{in},\z_{in})} -3\logp{\frac{\sum_{j=1}^4 p_j(\x_{in},\z_{in})}{4}}
\big].
\end{align*}
Now, due to the concavity of the function $log$, we have (by Jensen's inequality):
\begin{align*}
   &\sum_{j=1}^4 \logp{p_i} \leq 4 \logp{\frac{\sum_{j=1}^4 p_j(\x_{in},\z_{in})}{4}}\\
   &\implies \sum_{k\neq i}\logp{p_k(\x_{in},\z_{in})} -3\logp{\frac{\sum_{j=1}^4 p_j(\x_{in},\z_{in})}{4}} \\&\leq \logp{\frac{\sum_{j=1}^4 p_j(\x_{in},\z_{in})}{4}} - \logp{p_i(\x_{in},\z_{in})} \forall i \in {1,2,3,4}\\
   &\implies C(G,E) \leq -\logp{4^9}+ \mathbb{E}_{(\x_{in},\z_{in}) \sim p_1}\big[ \logp{\frac{\sum_{j=1}^4 p_j(\x_{in},\z_{in})}{4}} - \logp{p_1(\x_{in},\z_{in})} 
\big]
\\&+
\mathbb{E}_{(\x_{in},\z_{in}) \sim p_2}\big[
    \logp{\frac{\sum_{j=1}^4 p_j(\x_{in},\z_{in})}{4}} - \logp{p_2(\x_{in},\z_{in})}
\big]\\
&+\mathbb{E}_{(\x_{in},\z_{in}) \sim p_3}\big[
    \logp{\frac{\sum_{j=1}^4 p_j(\x_{in},\z_{in})}{4}} - \logp{p_3(\x_{in},\z_{in})}
\big]
\\&+
\mathbb{E}_{(\x_{in},\z_{in}) \sim p_4}\big[
    \logp{\frac{\sum_{j=1}^4 p_j(\x_{in},\z_{in})}{4}} - \logp{p_4(\x_{in},\z_{in})}
\big]\\
&\implies C(G,E) \leq-\logp{4^9} - KL\left(p_1||\frac{\sum_{j=1}^4 p_j(\x_{in},\z_{in})}{4}\right) - KL\left(p_2||\frac{\sum_{j=1}^4 p_j(\x_{in},\z_{in})}{4}\right) \\
&- KL\left(p_3||\frac{\sum_{j=1}^4 p_j(\x_{in},\z_{in})}{4}\right) - KL\left(p_1||\frac{\sum_{j=1}^4 p_j(\x_{in},\z_{in})}{4}\right)\\
&\implies C(G,E) \leq -\logp{4^9} - JSD_{\frac{1}{4},\frac{1}{4},\frac{1}{4},\frac{1}{4}}(p_1,p_2,p_3,p_4)
\end{align*}
Since each of the KL terms and correspondingly the JSD is non-negative, we have:
\begin{align*}
    C(G,E) \leq -\logp{4^9}
\end{align*}
Moreover, all the KL terms vanish simultaneously (and correspondingly the JSD) if and only if $p_1 = p_2 = p_3 = p_4$, leading to the global optimum given by:
\begin{align*}
    p_{(\x, E(\x))} = p_{(G(\z), \z)} = p_{(\x, E(G(E(\x))))} = p_{(G(E(G(\z))), \z)}
\end{align*}
Thus, both the original objective and the modified objective have the same global optimum.

\subsection{Ablations for Self-Supervised and Knowledge-based Feedback}
Results in table \ref{sample-table-2} demonstrate than even though original ALI model augmented with outputs of pre-trained inception net achieves improvements in reconstruction quality over the ALI baseline, the GALI-4 and GALI-pretrain models still significantly outperform it in pixel and feature level reconstruction quality. Similarly, the quantitative results in table table \ref{sample-table} demonstrate that the GALI-4 model without self supervision performs worse than GALI-mix on the image inpainting task. These results establish that it is beneficial to combine different types of feedback instead of using them in a stand-alone manner. 
\begin{table}[htb]
\caption{\small Pixel-wise Mean-Squared Error (MSE) and Feature level Mean Squared Error (MSE) on reconstruction task for CelebA. Lower is better.}
\label{sample-table-2}
\vskip 0.15in
\begin{center}
\begin{small}
\begin{sc}
\begin{tabular}{lcccr}
\toprule
Model  & pixel-MSE & feature-MSE\\
\midrule
ALI-pretrain & $0.068\pm
0.0053$ & $0.27 \pm 0.014$\\
ALI &  $0.074
 \pm 0.004$  & $0.307
 \pm 0.018$ \\
ALICE &  $0.042 \pm
0.002$ &
$0.248 \pm
0.013$ \\
GALI-4 & $\mathbf{0.036} \pm
0.002$ & $
\mathbf{0.201} \pm
0.012$ \\
GALI-PT &$\mathbf{0.032} \pm 0.0019$
& $\mathbf{0.131} \pm
0.008$\\
ALICE-PT &$\mathbf{0.053} \pm
0.003$ &$\mathbf{0.212} \pm 0.015$\\
\bottomrule
\end{tabular}
\end{sc}
\end{small}
\end{center}
\vskip -0.1in
\label{tb:reconst}
\vspace*{-0.3cm}
\end{table}
\\
Unlike L1/L2 based reconstruction of features, our model with knowledge-based feedback simultaneously matches the joint distributions of the output features of the pretrained model with all classes of images and encodings (fake, reconstructed and real). To verify the effectiveness of our approach over explicit reconstruction error based models, we conducted an ablation study against a model trained with an explicit $L_2$ reconstruction error in the feature space (using the same feature maps for both the datasets as used for GALI-PT).
We observed that the pre-trained features based $L-2$ regression error model (denoted by ALICE-PT in the above table) suffers from a trade-off similar to ALICE, improving upon ALI on metrics directly related to the pretrained features reconstruction error such as feature MSE (SVHN: 0.0724, CelebA: 0.212) while still performing poorly compared to GALI on other metrics such as pixel-level MSE (SVHN:0.0157, CelebA:0.053) and misclassification rate(SVHN:17.21).
\subsection{Architecture and Setup}

For all our proposed models and baselines, we utilize the architectures for encoder, generator and discriminator proposed in ALI as our base architecture with the discriminator modified to use Spectral Normalization instead of batch normalization and dropout. The ALI architecture can be divided into the following components: $G_z(x)(encoder),G_x(z)(generator),D(x),D(z)$ and $D(x,z)$. Here $D(x)$ and $D(z)$ convert image and latent vectors respectively to feature vectors using convolutional and fully-connected layers. These two feature vectors are subsequently concatenated and input to $D(x,z)$ which utilizes fully-connected layers to output the final output probability/probabilities.  To further improve stability, we add an exponentially decaying Gaussian noise to each image fed to the discriminator and use $tanh$ non-linearity for the generator's output layer. We found that these modifications were essential to reproduce the representation learning results reported in ALI. To keep the objectives and learning rates in similar range for all the models, we weigh an objective containing $m$ log terms by $2/m$. This corresponds to a weight of $1$ for the ALI's generator and discriminator objective. We use the same architecture for all our proposed models and ALI/ALICE baselines and further perform grid-search for the learning rate ($\alpha$), number of discriminator steps per generator steps($n_{dis}$), and the regularization parameter for ALICE in the range ($2\times10^{-5}, 4\times10^{-5}, 1\times10^{-4}, 2\times10^{-4}$),($1,2,5$), and $(10^{-4},10^{-2},1,10^{2})$ respectively. For all our experiments, we use Nvidia GeForce GTX 1080 Ti GPUs.

\subsection{Extension to Arbitrary Number and Size of Tuples}
For the GALI-8 model, the discriminator needs to be modified to input tuples of the form (image, latent-vector, image, latent-vector). To achieve this, we first input both images and both latent vectors through the $D(x)$ and $D(z)$ components of the discriminator respectively to output four feature vectors. These are subsequently concatenated and input to the discriminator's $D(x1,z1,x2,z2)$ component which is constructed by doubling the input size of the first fully connected layer in ALI's $D(x,z)$ component.

\subsection{Pretrained models}
Here we provide a brief description of the pretrained models used for calculating feature-MSE for svhn and celeba datasets. For CelebA, GALI-pretrain uses feature vectors from pre-trained inception v3 network. For all datasets, we perform bilinear interpolation whenever the input size of the pretrained model is different from the used data.
\begin{itemize}
    \item SVHN: The multi-digit detector used first passes each image through a shared component containing sequence of convolutional and fully connected layers to output a fixed length feature vector. This feature vector is then separately passed through different networks composed of fully connected layers which output the number of digits and the classification of each digit. For our experiments, we use the fixed length feature vector output from the shared component as it captures both the identity and length of the number present in the image.
    \item Celeba: For Celeba, we use the outputs of the last hidden layer of a pre-trained multi-label classifier composed of convolutional and fully connected layers trained to indentify all of the 40 attributes for the Celeba dataset.
\end{itemize}
\subsection{Image Generation Samples (CelebA)}
In Figure 56, we qualitatively compare the generation quality of our proposed models, baselines as well as the models based on the original mini-max and the misclassification likelihood objective. We observe that both mini-max and the misclassification likelihood objectives are unable to train the model to produce realistic images wheres the model trained on product of terms objective leads to comparable image generation quality to ALI while significantly improving the reconstruction quality. Moreover, our model does not suffer from the blurriness effect in generations and reconstructions visible in the results obtained for ALICE.
\begin{figure*}\begin{minipage}[b]{.3\textwidth}\centering \includegraphics[width=\textwidth]{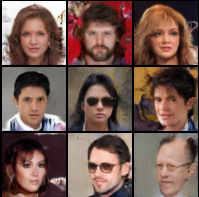}\subcaption{ALICE}\label{fig:mainfig}\end{minipage}%
\hfill
\begin{minipage}[b]{.3\textwidth}\includegraphics[width=\textwidth]{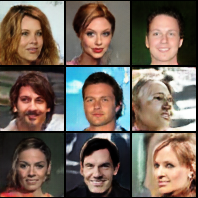}\subcaption{ALI}\label{fig:tuples}\end{minipage}
\hfill
\begin{minipage}[b]{.3\textwidth}\centering \includegraphics[width=\textwidth]{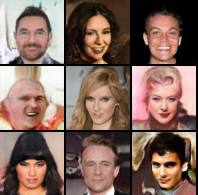}\subcaption{GALI-4}\label{fig:tuples}\end{minipage}

\begin{minipage}[b]{.3\textwidth}\centering \includegraphics[width=\textwidth]{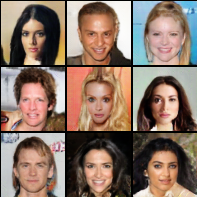}\subcaption{GALI-pretrain}\label{fig:mainfig}\end{minipage}%
\hfill
\begin{minipage}[b]{.3\textwidth}\centering \includegraphics[width=\textwidth]{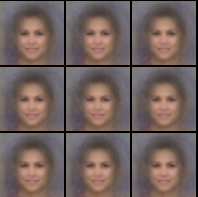}\subcaption{GALI-4 with minimax objective}\label{fig:tuples}\end{minipage}
\hfill
\begin{minipage}[b]{.3\textwidth}\centering \includegraphics[width=\textwidth]{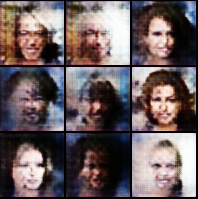}\subcaption{GALI-4 with misclassification likelihood objective}\label{fig:tuples}\end{minipage}

\caption{Qualitative results of generated samples for the CelebA dataset~\cite{celeba} from various versions of the proposed approach GALI and the baselines (ALI and ALICE). We did not include these generated samples in the main paper due to space constraint.}\label{fig:titlefig}
\vspace{-1em}
\end{figure*}

\subsection{Quantitative Evaluation for Image Inpainting}
Similar to the evalution of reconstruction quality, we quantitatively evaluate image in painting capabilities of the models using both pixel and feature level metric. In table \ref{sample-table}, pixel-level MSE denotes the average pixel wise squared difference of the inpainted region with the original image whereas the feature level MSE is calculated as the average squared difference between the feature vectors of the inpainted image and the original image for each model. The feature vectors are calculated using the same pretrained classifier as used for the reconstruction quality.In the table, we denote our proposed mixed images based model as GALI-mix while ALI and ALICE denote models obtained by utilizing the same distribution over masks as GALI-mix while feeding images to the encoder in the respective baselines. GALI-4 similarly denotes the GALI-4 model augmented with masked inputs. Results for GALI-4 are included as part of an ablation study.
\begin{table}[htb]
\caption{\small Pixel-wise Mean-Squared Error (MSE) and Feature level Mean Squared Error (MSE) on the image-inpainting task for CelebA. Lower is better.}
\label{sample-table}
\vskip 0.15in
\begin{center}
\begin{small}
\begin{sc}
\begin{tabular}{lcccr}
\toprule
Model  & pixel-MSE & feature-MSE\\
\midrule
ALI~\cite{ali} &  $0.060 \pm
0.0080$  & $0.38 \pm
0.016$ \\
ALICE~\cite{alice} &  $0.047 \pm 0.0069$ &
$0.32 \pm
0.015$\\
GALI-4 & $0.046 \pm
0.0066$ &
$0.285 \pm
0.012$\\
GALI-mix & $\mathbf{0.031} \pm 0.0050$
& $\mathbf{0.164} \pm 0.010$\\
\bottomrule
\end{tabular}
\end{sc}
\end{small}
\end{center}
\vskip -0.1in
\vspace*{-0.3cm}
\end{table}

\appendix

% This document was modified from the file originally made available by
% Pat Langley and Andrea Danyluk for ICML-2K. This version was created
% by Iain Murray in 2018, and modified by Alexandre Bouchard in
% 2019 and 2020. Previous contributors include Dan Roy, Lise Getoor and Tobias
% Scheffer, which was slightly modified from the 2010 version by
% Thorsten Joachims & Johannes Fuernkranz, slightly modified from the
% 2009 version by Kiri Wagstaff and Sam Roweis's 2008 version, which is
% slightly modified from Prasad Tadepalli's 2007 version which is a
% lightly changed version of the previous year's version by Andrew
% Moore, which was in turn edited from those of Kristian Kersting and
% Codrina Lauth. Alex Smola contributed to the algorithmic style files.

\end{document}